\documentclass[twoside]{article}

\usepackage{blindtext} 

\newcommand {\LiLT} {\textbf{L{\kern.01em}iLT}}

\usepackage{xcolor}
\usepackage{tikz}
\usetikzlibrary{arrows.meta}
\usetikzlibrary{positioning, fit, backgrounds}

\tikzset{
  var/.style={circle, draw, thick, minimum size=0.8cm, inner sep=2pt},
  plate/.style={draw, inner sep=10pt, rounded corners},
  lbl/.style={font=\small\itshape}
}

\tikzset{
  atom/.style={draw, rectangle, rounded corners=2pt, inner sep=3pt, font=\small},
  graybox/.style={atom, fill=gray!10},
  goldbox/.style={atom, fill=yellow!20, draw=yellow!50!black},
  group/.style={draw, dashed, rounded corners=5pt, inner sep=10pt},
  title/.style={font=\bfseries}
}

\tikzset{
  var/.style={
    rectangle,
    rounded corners=2pt,
    fill=gray!20,
    minimum size=0.8cm,
    inner sep=2pt
  },
  plate/.style={draw, inner sep=10pt, rounded corners},
  lbl/.style={font=\small\itshape}
}
\usepackage{amsmath}
\usepackage{amssymb}
\usepackage{algorithm}
\usepackage[noend]{algpseudocode}
\usepackage{hyperref}

\makeatletter
\algrenewcommand\ALG@beginalgorithmic{\small}

\algrenewcommand{\algorithmiccomment}[1]{%
  \hfill{\footnotesize\textcolor{gray}{$\triangleright$ \textit{#1}}}%
}
\makeatother

\usepackage[most]{tcolorbox} 
\usepackage{xfrac}

\usepackage{iftex}
\ifPDFTeX
  \usepackage[T1]{fontenc}
  \usepackage[utf8]{inputenc}
  \usepackage{tgtermes} 
  \usepackage{newtxmath} 
  
\else
  \usepackage{fontspec}
  \newfontfamily\ipafont{Doulos SIL}
  
\fi

\newcommand{\Sref}[1]{\S\ref{#1}}

\usepackage{makecell}
\usepackage{multirow}
\usepackage{lscape}
\usepackage{setspace}
\usepackage{mathtools} 

\usepackage[main=english]{babel}

\usepackage{float} 
\usepackage{tabularx} 
\usepackage[hmarginratio=1:1,top=1.25in,columnsep=20pt,papersize={6in,9in}]{geometry} 
\usepackage[hang, small,labelfont=bf,up,textfont=it,up]{caption} 
\usepackage{booktabs} 

\usepackage{enumitem} 
\setlist[itemize]{noitemsep} 

\usepackage{abstract} 

\usepackage{titlesec} 

\usepackage{fancyhdr} 
\fancypagestyle{firststyle} 
{
   \fancyhf{}
   \fancyfoot[C]{\thepage}

}

\usepackage{titling} 

\usepackage{natbib}

\usepackage{gb4e} 
\noautomath

\makeatletter
\def\blfootnote{\xdef\@thefnmark{}\@footnotetext}
\makeatother

\title{\bfseries{Task Decomposition for Efficient Annotation}} 
\author{%
\textsc{Nupoor Gandhi, Emma Strubell} \\[1ex] 
\normalsize \textit{Carnegie Mellon University} \\ 
}
\date{} 
\renewcommand{\maketitlehookd}{%
\begin{abstract}
\noindent High-quality annotations of structured representations are expensive to collect over large corpora. 
Manual annotation of structure is laborious, and model-based annotation, although cheaper to generate, requires expensive validation and potentially significant supervision to ensure that the annotation quality is strong enough to be useful downstream.
In traditional annotation workflows, annotation of each complete example is performed end-to-end by a single annotator. 
However, structured annotation is complex, and each aspect of the task represents a unique challenge with an associated inferential load for a given annotator. 
Modern annotation projects can incorporate heterogeneous groups of annotators, including both models and human annotators with varying domain and linguistic expertise. 
It remains unclear, however, how to redesign annotation tasks in this setting, where efforts are discriminately allocated across heterogeneous annotators with respect to distinct annotation challenges.

We propose to decompose annotation tasks into sub-tasks in order to reduce the aggregate inferential load of annotation projects. Inspired by the notion of centers from centering theory, we introduce a formal model of inferential load based on the degrees of freedom in the space of valid annotations. 
Using this model, we show that identifying these centers (i.e. salient anchor entities realized by annotation sub-tasks) constrains the output space complexity, and decompositions which isolate and advance center identification reduce the aggregate inferential load.
We provide guidelines for decomposing complex structured annotation tasks, supported by examples demonstrating improved cost-efficiency from our prior work.
Finally, we present a procedure for allocating sub-tasks across annotators to maximize quality under a fixed budget.
\end{abstract}
}

\begin{document}
\tolerance 1414
\hbadness 1414
\emergencystretch 1.5em
\hfuzz 0.3pt
\widowpenalty=10000
\vfuzz \hfuzz
\raggedbottom

\maketitle
\thispagestyle{firststyle}


\section{Introduction}

Structured representations of text persist as a valuable modality for domain experts analyzing large corpora, where \textit{structuring} refers to a mapping from text to a semantic, graphical representation of meaning composed of span-bound atomic units and relations, where the particular structure may be selected to support an application \citep{abend-rappoport-2017-state}. Structure is necessarily grounded in text, which affords reliability in contrast with automating analyses completely \citep{maynez-etal-2020-faithfulness, ijcai2025p873}, and structure is a compact, natural representation of the global properties of entities that experts value. 
Further, projection of structure onto text is a necessary step in inductive coding, which remains the standard for reliable qualitative methods \citep{glaser2017discovery}.
Rich structured annotation is historically prohibitively expensive to develop over large volumes of text.
This is especially true of specialized domains \citep{kim2003genia, li2016biocreative, krallinger2015chemdner}.
It took thousands of hours to annotate concepts for the CRAFT biomedical article dataset \citep{bada2012concept}.
For the POLIANNA policy design dataset, \citet{sewerin2023towards} report over 600 hours of annotation. 
The problem is compounded for high-dimensional, ambiguous tagsets or cases where documents are long and require expertise to interpret \citep{wright-bettner-etal-2020-defining}. 

As LLMs and, more generally, neural architectures have improved in annotation capabilities, a natural approach to reducing costs is to substitute human annotators with models.
In the conventional annotation paradigm, after agreement between annotators is established, each example is annotated in its entirety by individual annotators. 
To incorporate models as annotators, this paradigm can be adapted where models may substitute humans, potentially focusing on the subset of unambiguous examples \citep{li-etal-2023-coannotating}. 
Humans optionally annotate training examples which are used to develop the model \citep{rebholz-schuhmann-etal-2010-calbc, hahn-etal-2010-proposal}. 
Alternatively, LLMs pre-annotate for the complete task, and human annotators optionally refine the pre-annotations \citep{naraki2024augmenting, Goel2023LLMsAA}. 
Regardless of which annotation process is selected, human annotation effort is necessary for validation of the process.

While LLM-generated datasets such as SynthIE \citep{josifoski2023exploiting}, PileNER \citep{zhou2023universalner}, and NuNER \citep{bogdanov-etal-2024-nuner} are widely used, naive substitution of LLMs for human annotators does not necessarily transfer to extraction over specialized domains representing a conceptual distribution shift \citep{zhang-etal-2024-scier, golde-etal-2025-familarity}, especially with respect to annotation specifications that require a substantial number of examples to robustly adhere to, such as span boundary rules \citep{volkanovska-2025-large}. 
Such tasks exhibit high \textit{inferential load}, with a vast space of valid annotations, where annotators make several decisions with associated dependencies in a single pass, some of which may require domain or linguistic expertise. 
Any limiting aspect of the task will bottleneck model performance.
Accordingly, developing high-performing model-based annotation for specialized domains can be costly, making use of hundreds of annotated examples for finetuning \citep{zhou2023universalner, sainz2023gollie}. 
%

In this work, we propose to decompose tasks into sub-tasks that reflect distinct aspects of annotation complexity in order to more efficiently spend scarce annotation resources. 
In a heterogeneous, multi-annotator setting including both human and model annotators, we argue for (1) the re-design of annotation tasks to reduce the aggregate inferential load, and (2) the redistribution of annotator efforts to aspects of a task that can be performed efficiently for the given annotator. 
First, to reduce aggregate inferential load, decomposition of annotation tasks can isolate the identification of centers—salient discourse entities that are tracked and updated over the course of annotation, as characterized by \citep{grosz1995centering}—since jointly inferring centers and their attributes expands the effective output space, whereas fixing centers apriori collapses this space and simplifies subsequent decisions.
Second, decomposition allows for the redistribution of annotator efforts such that each atomic sub-task can be assigned to the least costly, adequate annotator, and expert effort can be reserved for the most demanding aspects of a task.
We first specify the desiderata for annotation projects in heterogeneous multi-annotator settings (\Sref{sec:desiderata}).
Our primary contribution is a model of inferential load for structured representation annotation (\Sref{sec:framework}) supporting task decomposition as a more efficient use of scarce annotation resources. 
With example annotation tasks, we illustrate the savings in aggregate inferential load from decomposition, which we support by highlighting empirical evidence from our own recent work that task decomposition is a more efficient use of annotation resources (\Sref{sec:examples}).
For brevity, we present only an overview of core results from our supporting works, and we encourage interested readers to refer to the original publications for further details.
We additionally provide a procedure for allocating annotation sub-tasks to diverse sets of annotators such that the budget is satisfied and annotation quality is maximized (\Sref{sec:procedure}).

\section{Design Objectives for Annotation Projects in Multi-Annotator Settings}\label{sec:desiderata}

We begin by reviewing the core goals of annotation projects in a setting with a heterogeneous set of annotators.

\paragraph{Low Cost} The cost of an annotation project depends both on the number of examples that need to be annotated and the inferential load required for each individual example. 
The inferential load of annotation depends on the complexity of an annotation structure, which is tied to utility, highlighting a trade-off between inferential load and downstream utility. 
For example, a large, granular set of ambiguous, higher order tags may be more useful than a small, less expressive set of simple tags, but at the same time exert a higher cognitive load on the annotator \citep{ma-etal-2023-large}.

For a given task, the inferential load for a human or model annotator is not necessarily the same, in absolute terms or relative to other tasks. 
For a given task, we can also expect the cost of model inference and training, if it is performed, to differ from the cost of human expert annotation. 
We can expect that for heterogeneous sets of annotators including humans and models, the relative costs of tasks will differ. 
To spend annotation resources efficiently, a given annotator should be assigned such that their efforts are concentrated on the aspects of a task that are cheap relative to other annotators.

\paragraph{High Quality} In addition to minimizing cost, it is necessary to establish both reliability, in that the task is well-defined and can be annotated consistently, and validity, in that that the task captures the truth about the phenomena we are interested in \citep{artstein-poesio-2008-survey}. 
Since annotators can share the same prejudices, strong agreement does not necessarily imply validity, so agreement is a necessary but insufficient condition. 
For example, \citet{bayerl-paul-2011-determines} finds that expert annotators and non-expert annotators tend to have comparable agreement within-group, and poorer agreement for a mixed group of experts and non-experts.
In practice, however, high agreement with an expert or consensus annotation is a widely-used heuristic for annotation quality.

To show that the task is well-defined, consistency can be established between annotators over a sample of data.
Consistency is a heuristic for shared understanding of guidelines across annotators, which is necessary to demonstrate that the task is well-defined, and that the proposed annotation procedure captures the purported phenomena documented in the annotation guidelines.
Standard measurements of consistency must be adapted in a heterogeneous, multi-annotator setting. Consistency between a pair of model annotators is not necessarily meaningful. 
For example, model annotators may share a similar training distribution, resulting in common biases and artificially high agreement 
\citep{10.1145/3708359.3712091, bavaresco-etal-2025-llms, lu-etal-2025-llm} with collapsed diversity in judgment \citep{messeri2024artificial}. 
Agreement between a machine and human annotator is also not without limitations, as high agreement can suggest that the task is shallow or surface-driven and not necessarily well-defined. Historically, agreement between human experts is the standard approach to establish the clarity of annotation guidelines \citep{pustejovsky2012natural}. 
Regardless of how agreement is measured, more informative signals may come from component-level consistency, since different parts of a task vary in their importance for downstream utility and in the extent to which they introduce disagreement with respect to the complete task, as seen in CRAFT concept annotation where unitization is a major bottleneck in annotator training \citep{bada2012concept}.


The extent to which task decomposition can reduce validation labor remains an open question that we leave for future investigation. In this work, we propose decomposition primarily as a vehicle for reducing inferential load and efficient distribution of annotation efforts.



\section{Framework}\label{sec:framework}
Necessary steps to reduce the cost of high-quality annotation projects include minimizing the aggregate inferential load and efficiently allocating annotator efforts. In this work, we argue that decompositions can produce both these outcomes.

First, with respect to aggregate inferential load, we will observe that the identification of \textit{centers}, salient discourse entities that are realized over the course of annotation, is a primary source of complexity in structured annotation. 
Drawing on centering theory \citep{grosz1995centering}, which characterizes coherent discourse in terms of continuous centers and minimal shifts in attention, we model annotation as a process that requires tracking and updating such entities. 
When centers are not specified in advance, annotators must simultaneously identify and characterize them, along with any dependencies prescribed by the structure. 
This joint burden expands the effective output space, as the annotator considers each plausible center assignment alongside its set of attributes. 
In contrast, when centers are fixed, the output space collapses and subsequent decisions are constrained. In a decomposed setting, we illustrate that the aggregate output complexity can be reduced by identifying centers in isolation of other aspects of the annotation task.


Second, task decomposition allows for more efficient allocation of annotator effort, where sub-tasks are assigned to the least costly, adequate annotator. 
Sub-tasks represent focused, atomic tasks, with strictly fewer inferential leaps than the complete task.
Under decomposition, impeding aspects of annotation can be isolated as sub-tasks, such that expensive annotator expertise is spent only on the aspects of the task where it is necessary.

To analyze how annotation design shapes resource efficiency, we formalize the annotation process in a way that makes inferential load and effort allocation explicit. 
This framework enables a precise characterization of the annotation project design problem (\Sref{sec:problem-formulation}), a model for quantifying inferential load (\Sref{sec:inf-load-model}), and a concrete definition of task decomposition (\Sref{sec:decomposition}).

\subsection{Problem Formulation}\label{sec:problem-formulation}

Our objective is to design an annotation project that produces annotations of the highest possible quality over the complete corpus given an annotation task, a set of annotators, and a fixed budget.
In designing an annotation task, we refer to two core mapping decisions. The first decision, is if and how to \textit{decompose} the annotation task $t$ into sub-tasks $t_1, t_2, \ldots, t_k$, where executing the sub-tasks is equivalent to executing the complete task. In the second decision, annotators $A_1, \ldots, A_{\ell}$ are assigned to sub-tasks, where each sub-task is performed by one annotator. 
In this work, the decomposition and an annotator assignment is selected to maximize annotation quality, which we take to be agreement with expert human annotation of a sample of the corpus.



\subsection{Model of Inferential Load} \label{sec:inf-load-model}

In this section, we propose to model inferential load as a function of three components: (1) the prerequisite information processing necessary to perform the annotation, (2) the space of decisions that the annotator considers, and (3) the quantity of salient entities realized by the annotation.
The necessary information processing for annotation includes the document to be annotated and the task instructions, which we formalize as sets of span-bound units.
The space of all possible annotations refers to the broad set of valid annotations given the parameters of a task, of which the annotator selects one. 
Finally, the process of annotation introduces salient entities from the task input, representing slots that the annotation task instructions expect to be filled by the output of annotation.
To more formally measure these aspects of inferential load, we develop a framework to analyze annotation in its complete or decomposed form.
\begin{figure}[h]
\centering
\includegraphics[width=\linewidth]{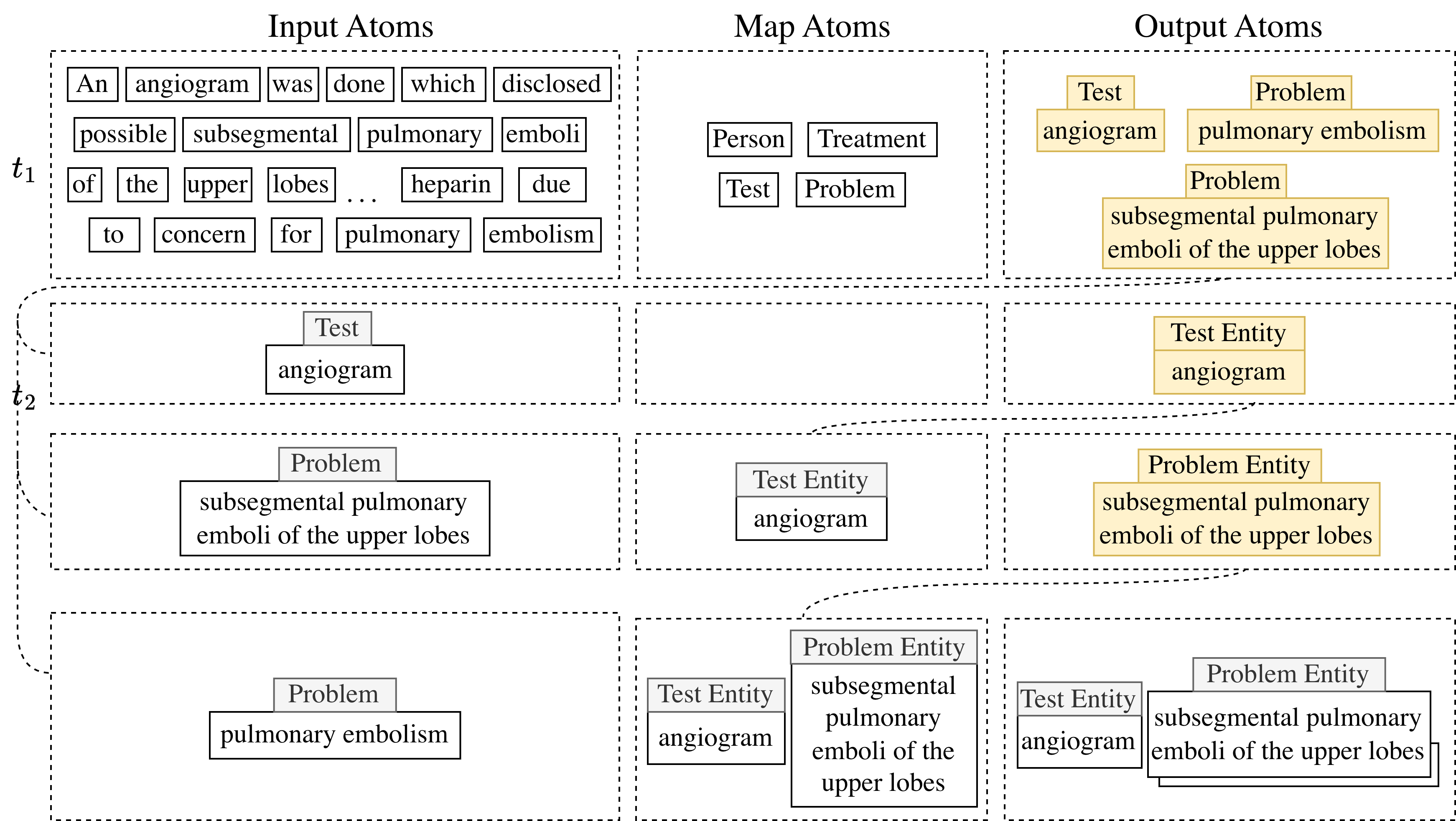}
\caption{We illustrate the atomic space for a coreference resolution task decomposition, where $t_1$ is typed mention identification and $t_2$ is antecedent linking. The top row represents the atomic space of a single instance of task $t_1$, and the subsequent rows represent instances of the task $t_2$. The dashed links between boxes denote the transfer of seed centers between sub-tasks. We denote the introduction of a new center in \textcolor[HTML]{d6b656}{yellow}. Each extracted mention represents a new center, and centers are introduced for each new entity created at the linking sub-task. See the graphical representation of this decomposition of coreference resolution in \autoref{fig:example-coref}.}
\label{fig:atomic-space}
\end{figure}
\subsubsection{Annotation Task}
An annotation task is a transformation of a structured or unstructured input to a structured output using a set of instructions. More formally, an annotation task instance is a function that transforms a set of input atoms, or markable units, to a set of output atoms, or annotated units, using a set of map atoms, or mapping concepts. An atom represents a sequence of tokens or a sub-structure, with optional properties or characteristics. We define the three types of atoms below.
\begin{enumerate}
    \item \textit{Map atoms}: Representing the structure the document is mapped to, map atoms appear in the task instructions (e.g. class types for NER, relation types for relation extraction, event triggers for event argument extraction). Map atoms may also be introduced from the input atoms while performing the task (e.g. a new active entity in coreference resolution).
    \item \textit{Input atoms}: The units to map primarily appear in the document (e.g. mentions for mention identification). The input atoms are restricted to the document context necessary to perform the annotation task, not necessarily the complete original input document. 
    \item \textit{Output atoms}: The atoms representing the product of annotation (e.g. typed mentions, a pair of typed mentions and a relation type, linked coreferent mentions with a corresponding entity type).
\end{enumerate}
\autoref{fig:atomic-space} represents an example atomic space for a decomposed coreference resolution task.
We additionally define two primitive annotation tasks in terms of these atomic spaces, where a task is a sequence of one or more of the following primitives performed in secession.
\begin{enumerate}
    \item \textit{Atom selection}: One or more subsets of contiguous atoms are selected from the set of input atoms.
    \item \textit{Atom classification}: One or more subsets of atoms are classified from the input atoms with respect to one or more map atoms.
\end{enumerate}

We also introduce \textit{centers}, which are the entities salient to the annotation task appearing in both the annotated output atoms and the input atoms. Centers are the discourse entities realized during annotation, where the entities represent individuals, objects, or concepts that can be tracked over the course of annotation.
In other words, centers represent the entities that annotators are tasked to identify or characterize.
An annotation task may specify that entities are identified from an atomic space, in which case the resultant output atoms represent new centers (\textit{center identification task}).
Otherwise, the annotation task specifies that information about an entity or group of entities be updated, in which cases, the current set of centers is enriched (\textit{enrichment task}). 
This may entail marking the precise span boundaries for a given mention vicinity, classifying the entity type of mention, or determining if the mention is referent with another mention.
This notion of centers in part follows from centering theory \citep{grosz1995centering} and focus spaces \citep{grosz-sidner-1986-attention}, where each annotation task instance is associated with a unique focus space containing salient entities. 
In our case, salient entities are those which are mentioned explicitly in the annotation output or are necessary for performing the annotation.
Centers are not bound to a precise surface form, and we expect that they evolve in atomic representation over the course of annotation. The notion of a center is more meaningfully fixed in the annotator's mental model, where the centers are iteratively updated with new information for chains of dependent sub-tasks. 
Rather than consider a single center at a time, our model assumes that annotators consider multiple centers at a time, where there is an inferential cost incurred from introducing a new center to the set, signaling a shift in focus \citep{grosz-sidner-1986-attention}.


\subsubsection{Features of Annotation Tasks Affecting Inferential Load}

\paragraph{Input and Map Space}
Given the input atoms and map atoms, the fixed dimensionality of the atomic spaces can be computed. The input space represents the relevant context that the annotator must process in order to perform the annotation. The map space represents the active sub-structure that the input atoms can be projected onto, so the annotator necessarily considers the map space dimensionality. Accordingly, we assume both the input and map atomic space affect inferential load, as they represent the space of units and tools for transformation performed during annotation. 
\paragraph{Output Space} The output atomic space represents the set of all possible output atom sequences that are valid given the annotation task and input text. 
The dimensionality of this space can be computed using the length of the input atom sequence, the dimensionality of the tag set, and the task instructions. We provide examples for how to compute the dimensionality of the output space for a few tasks in \Sref{sec:examples}.

\paragraph{Center Introduction} 
In atom selection tasks, the surface form of centers are selected, and in atom classification tasks, the surface form of centers are classified.
We assume that for a given annotation task $t_{i}$, if there is a dependency between $t_{i-1}$ and $t_i$, then any number of centers from the output space of $t_{i-1}$ can be seeded as an anchor point for the annotation. 
Seed centers of an instance of task $t_i$, in this context, refers to any center that is passed as a parameter to the instance. 
We can see an example of mentions and entities passed as seed centers in \autoref{fig:atomic-space}.
In this work, we are especially interested in the quantity of centers that are introduced during the annotation task, where these new centers are disjoint from the seed centers, if there are seed centers. 

\subsection{Decomposition}\label{sec:decomposition}
An annotation task $t$ can be broken into a sequence of primitives $t_1, \ldots, t_k$. We model dependencies between primitives in a graphical model with plate notation, where executing the primitives in the order of a topological sort is equivalent to executing the task $t$. 
The graphical model represents the annotation of a single task, and each node in the graph represents the execution of an instance of the task. 
Edges between nodes $t_i$ and $t_{i+1}$ represents a dependency, such that the execution of task $t_{i+1}$ is dependent on the outcome of executing task $t_i$. 
For root tasks, the input atoms are from the initial unstructured input document (i.e. the input atoms for $t_0$). For each child task $t_{i+1}$, the input atom set is a function of the parent task $t_i$ output atoms and the initial input document. 
The map atoms are specified by the task $t_i$, and the output atoms of $t_i$ is the transformation of the input atom sequence. 
If there is an edge from $t_i$ to $t_{i+1}$, then any subset of centers can be passed as seed centers.

The plate notation represents repeating variables in the graphical model. If an edge from $t_i$ to $t_{i+1}$ crosses a plate, then the set of input atoms of the $j$'th repetition of the task $t_{i+1}$ is a function of the $j$'th output atom for $t_i$. 
The number of repetitions is specified in the bottom right corner of each plate.

\begin{figure}[h]
\centering
\includegraphics[width=\linewidth]{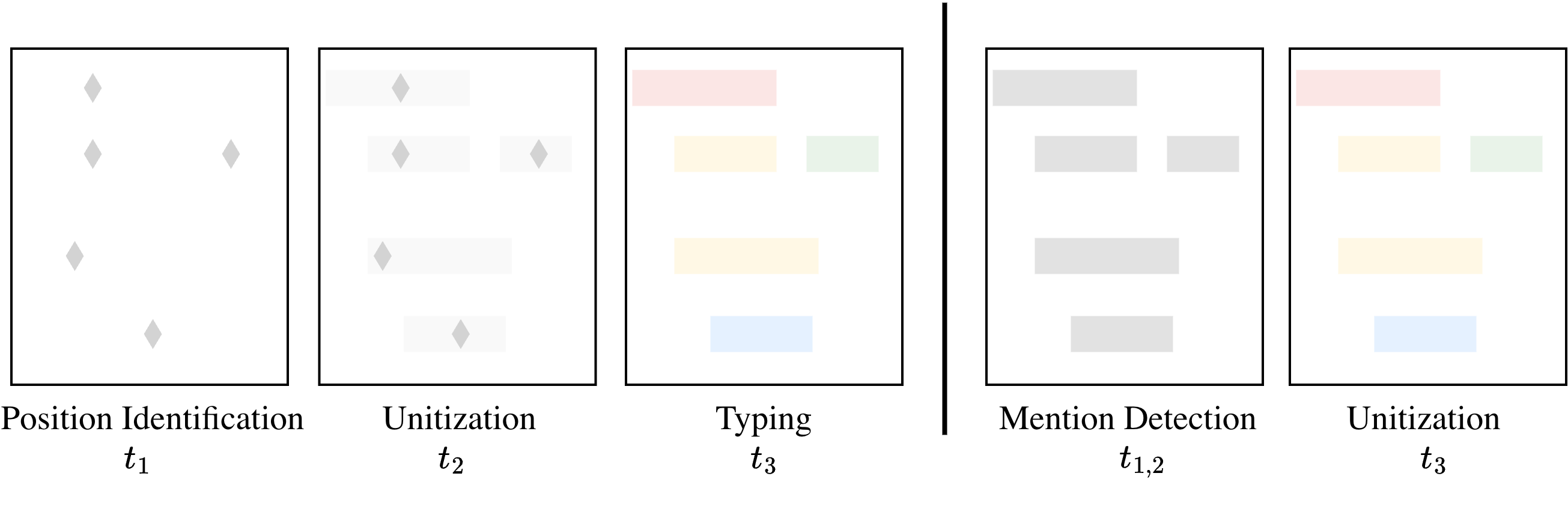}
\vspace{0.5em}
\begin{tikzpicture}[node distance=0.9cm, >={Stealth}]

  \node[var] (A) {$t_1$};
  \node[var, right=of A] (B) {$t_2$};
  \node[var, right=of B] (C) {$t_3$};

  \draw[->] (A) -- (B);
  \draw[->] (B) -- (C);

  \begin{pgfonlayer}{background}
    \node[plate, fit=(B)(C)] (p1) {};
  \end{pgfonlayer}
    \node[font=\footnotesize] at (p1.south east) [anchor=south east, inner sep=2pt] {$M$};

  \node[var, right=2cm of C] (D) {$t_{1,2}$};
  \node[var, right=of D] (E) {$t_3$};

  \draw[->] (D) -- (E);

  \begin{pgfonlayer}{background}
    \node[plate, fit=(E)] (p2) {};
  \end{pgfonlayer}
    \node[font=\footnotesize] at (p2.south east) [anchor=south east, inner sep=2pt] {$M$};

\end{tikzpicture}
\caption{The top row is a pictoral representation of two candidate decompositions of NER: $\{t_1, t_2, t_3\}$ and $\{t_{1,2}, t_3\}$. The bottom row is a graphical representation of the decompositions. In the primitive decomposition graph on the left, $t_1$ is position identification, $t_2$ is unitization, and $t_3$ is typing. Contracting the edge $(t_1, t_2)$ produces the graph on the right where $t_{1,2}$ represents mention detection. Both graphs are valid decompositions for the NER task $t$. In this example $t_1, t_2, t_{1,2}$ are atom selection tasks, and $t3$ is an atom classification task.}
\label{fig:ner-example-graphs}
\end{figure}
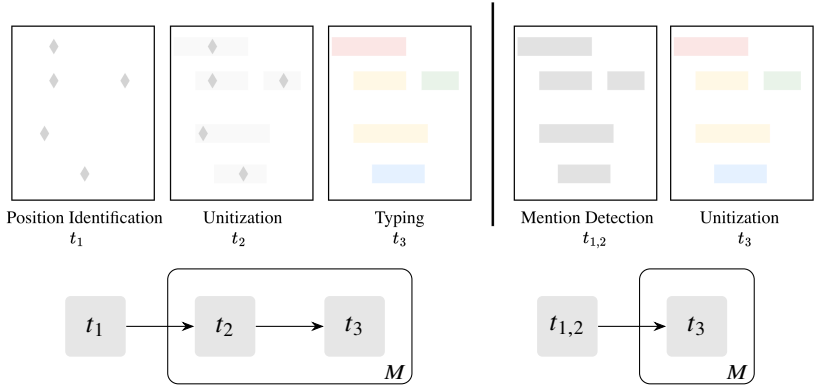

Applying any series of edge contractions to the primitive graphical model produces a valid decomposition. We can see examples of two valid decompositions of an NER task in \autoref{fig:ner-example-graphs}.

\subsubsection{Constraints on Decomposition}
A task should be decomposed if the decomposition represents a lower aggregate inferential load. 
We consider several dimensions of inferential load of an annotation task, specifically the combinatorial atomic output space, the input space, the map space, and the number of centers realized by the annotation task.
For an annotator $A$, we assume that each dimension has a unique impact on the annotator's inferential load, which we represent with weights $w_{\text{o}}, w_{\text{i}}, w_{\text{m}}$ and $w_{\text{center}}$. This gives us the following expression to represent inferential load for a decomposition $t_1, \ldots, t_k$.




\begin{equation}
\textsc{Inferential Load}(\{t_1, \ldots, t_k\}) =
\sum_{j=1}^{k} 
\begin{aligned}[t]
  &\textsc{Centers Introduced}(t_j) \cdot w_{\text{center}} \\
  &+ \textsc{Output Space}(t_j) \cdot w_{\text{o}} \\
   &+ \textsc{Input Space}(t_j) \cdot w_{\text{i}} \\
  &+ \textsc{Map Space}(t_j) \cdot w_{\text{m}}
\end{aligned}
\label{eq:inf-load}
\end{equation}
where across all instances of task $t_j$, \textsc{\# Centers Introduced}($t_j$) represents the number of new centers realized during annotation, $\textsc{Output Space}(t_j)$ represents the dimension of the combinatorial atomic output space, $\textsc{Input Space}(t_j)$ represents the number of atoms in the input spaces, and $\textsc{Map Space}(t_j)$ represents the number of atoms in the map spaces.
We constrain decompositions based on our notion of inferential load such that a valid decomposition should be used if \autoref{eq:constraint} is satisfied.
\begin{align}
    \textsc{Inferential Load}(\{t_1, t_2, \ldots, t_k\}) &\leq \textsc{Inferential Load}(\{t\})\label{eq:constraint}
\end{align} 

These dimensions of inferential load are intended to capture the information processing cost, the complexity of the annotation task, and finally the entities realized by annotation. 
The input and map space make up the units we consider when performing annotation, while the output space complexity represents the number of ways the annotation could be performed.
The number of centers is the number of meaningful new discourse entities in the annotation output.

\paragraph{Note on the conditions where \autoref{eq:constraint} is satisfied} 
Decomposition can reduce aggregate inferential load by reducing the output space complexity.
As we will observe in \Sref{sec:discussion}, the output space dimensionality shrinks when tasks are decomposed such that center introduction is isolated from enrichment tasks.
The other dimensions of inferential load may incur a larger cost under decomposition due to redundant information processing.
With respect to center identification, if a decomposition does not have a sequence of redundant or unnecessary sub-tasks, the decomposition will not introduce any more centers than the complete task. 
Task decomposition, does however create multiple instances of the same task, producing redundant information processing. Context will be reused to perform different sub-tasks in a decomposition, so the cost incurred by the aggregate input and map space may outweigh the benefit of reducing the output space complexity for some annotators. In these cases, decomposition would not reduce the inferential load, and the constraint would not be met. We elaborate further on the desiderata of decompositions that meet this constraint (\autoref{eq:constraint}) in subsequent sections \Sref{sec:examples} and \Sref{sec:discussion}.


\subsubsection{Relationship between Atomic Space and Inferential Load}\label{sec:relation-space-inf-load}

For machine annotation, decomposing tasks to reduce the aggregate output space is common and in some cases has been shown to be better performing.
\citet{zhou2023universalner} show that extracting mentions for individual types results in better performance than eliciting mentions for all types in a single pass.
\citet{liu-etal-2025-towards-event} decompose event extraction task into sub-tasks that are either performed with distant supervision or using an LLM annotator.
When extracting information from tabular data, it is common to decompose tasks into sub-tasks \citep{zhang-etal-2025-exploring} with intermediate representations of tabular data \citep{wang2024chain}.
To address the large pool of candidates for cross-sentence arguments in implicit event argument detection, \citet{zhang-etal-2020-two} propose to decompose the search to first detect the argument head word, followed by head-to-span expansion.
\citet{zhang-etal-2020-two} motivate this decomposition in the reduction of the output space dimension: by identifying head words first, the space of candidate spans scales linearly with the document length, not quadratically.

Decomposing tasks into sub-tasks with a smaller output space can make smaller models competitive with larger models.
\citet{bhattacharyya-etal-2025-information} finds that for information extraction, processing semantic blocks in visually rich documents independently results in better performance than processing whole documents in a single pass. They further find that the decomposition of processing can make smaller language models (32B) competitive with or outperform larger language models (200B). Similarly, \citet{laurer2023building} develop a BERT-based 0-shot classifier, decomposing a classification task into natural language inference tasks for individual classes, where each class is expressed as a hypothesis.

For human annotation, previous empirical and theoretical work suggests that the inferential load is lower for sub-tasks with a smaller atomic space.
\citet{wei2018clinical} investigate factors associated with NER annotation time for clinical text, finding that a larger atomic space is associated with higher annotation time (e.g. number of words, dependency distance, number of entities).
\citet{fort-etal-2012-modeling} establish several dimensions of annotation complexity, many of which underscore the high inferential load induced by a large atomic space.
For example, they put forth the dimension of discrimination of markable units from the complete input document, which is a function of the number of input atoms and output atoms.
The tag set dimension is a function of the number of map atoms.
The weight of the context for performing the annotation task is a function of the number of input atoms.
We note that some dimensions of complexity proposed in \citet{fort-etal-2012-modeling} are not captured by the size of the atomic space, namely the expressiveness and the degree of ambiguity of map or input atoms.

\subsubsection{Relationship between Center Introduction and Inferential Load}\label{sec:relation-centers-inf-load}

With respect to machine annotation, models tend to excel at center enrichment tasks, where the salient entities are pre-defined.
Providing gold triggers enables an unsupervised event extraction model to perform comparably to a fully supervised system trained on thousands of labeled examples \citep{zhang-etal-2021-zero}.
In coreference, LLMs can perform mention clustering robustly given gold mentions \citep{le2023large,sundar-etal-2024-major}.
In cross-document Abstract Meaning Representation parsing, \citet{ahmed-etal-2024-x} develop a co-pilot annotation system where given roleset IDs and event triggers, the model suggests arguments. They find that LLMs can accurately recover trigger lemmas and roleset IDs, while struggling with full argument structures.

Further, \citet{wei2018clinical} suggest that center identification may be a more important dimension of complexity than atomic space for manual annotation.
\citet{wei2018clinical} find that the number of entities appearing in a document has a larger effect on annotation time than the number of words in the document. Here, the number of entities can represent the minimum number of times that the focus shifts, and the size of the input atom space can be taken as the document length.


\section{Example Task Decompositions}\label{sec:examples}

We provide examples of task decompositions for two common annotation tasks: NER and coreference resolution to illustrate cases of valid decompositions which do and do not satisfy the constraint that decompositions reduce the aggregate inferential load (\autoref{eq:constraint}). 
We describe for each task the inferential load incurred by the decomposition in contrast with that of the complete task. 
We show that for the first two examples, applying our model for inferential load suggests that the decomposition would result in a lower aggregate inferential load, assuming a high relative cost for output space complexity. 
The low inferential load of decomposition in these examples is also supported by our previous empirical work where we validate efficiency gains from decomposition with respect to annotator time. 
For the last example, we find that decomposition results in a higher aggregate inferential load where there is unnecessary factorization and a high information processing cost ($w_{\text{i}}$, $w_{\text{m}}$) relative to output complexity ($w_\text{o})$.
\begin{table}
\centering
\footnotesize
\begin{tabularx}{0.65\textwidth}{@{}cX@{}}
\toprule
\text{Variable} & \text{Description} \\
\midrule
$N$ & Number of tokens in the document \\
$C$ & Number of characters in the document \\
$M$ & Number of mentions in the document \\
$S$ & Number of possible spans $\left(\text{i.e. }\binom{N+1}{2}\right)$ \\
$T$ & Number of entity types \\
$d$ & Maximum span width \\
$k$ & Number of entities  \\
$m_j$ & The $j$-th mention in a document \\
\bottomrule
\end{tabularx}
\caption{Variables used in task decomposition examples throughout \Sref{sec:examples}. We express the atomic space and the number of centers introduced as a function of these variables.}
\end{table}
\paragraph{NER} For NER, the complete task $t$ entails identifying named entity spans from a document. 
We can decompose the task into first selecting typed positions in the text where a span appears $t_1$, then unitizing the span, or identifying the span boundaries for the given typed position $t_2$ (\autoref{fig:example-ner}). 
We can observe a dependence between $t_1$ and $t_2$, where the position to be typed realized by $t_1$ is taken as a seed center in the input space of $t_2$. 
Each position identified in $t_1$ is typed and unified, so there are $M$ instances of $t_2$, where $M$ denotes the number of mentions in the document. 
If we let $N$ denote the number of tokens in the document, $S=\binom{N+1}{2}$ is the set of all possible spans. 
Given $T$ entity types, we have $(T+1)^S$ ways to assign types to the set of all possible spans, which is the atomic output space dimension for the complete task $t$. 
For the decomposition, there are $(T+1)^C$ possible typed position selections for a document with $C$ characters yielding the output complexity of $t_1$.
For a given position, if $d$ represents the number of characters in the longest span, there are at most $\binom{d+1}{2}$ ways to unitize the position.
The aggregate output complexity for the decomposition is $(T+1)^C + M \binom{d+1}{2}$, which is smaller than $(T+1)^S$ as we show in \Sref{sec:proof-ner}. This in combination with no additional centers gives us a lower aggregate inferential load for decomposition.

The value of decomposition for NER is supported in \citet{gandhi-bada-strubell}: We find that decomposing the concept annotation task to typed position identification and unitization results in a more efficient use of manual annotation time, where manual annotation is focused on position identification rather than unitization. The output complexity for typed position annotation is $(T+1)^C$ which is much smaller than the complete task $(T+1)^S$, suggesting that annotating typed positions instead of complete NER would take less time, which is supported by timed annotation experiments \citep{andrade2024boundary}. 
We also found unitization to be a cheap task to develop a machine annotator for, requiring as few as 100-200 example sentences, which is supported by our model of inferential load, as $t_3$ has a negligible atomic space with a small output complexity, and it introduces no new centers.

\begin{figure}[h]
\centering
\begin{tikzpicture}[node distance=0.9cm, >={Stealth}]
  \node[var] (A1) {$t_1$};
  \node[var, right=of A1] (B1) {$t_2$};
  \draw[->] (A1) -- (B1);
  \begin{pgfonlayer}{background}
    \node[plate, fit=(B1)] (p1) {};
  \end{pgfonlayer}
  \node[font=\footnotesize] at (p1.south east) [anchor=south east, inner sep=2pt] {$M$};
\end{tikzpicture}

\vspace{0.5cm}

{\footnotesize
\begin{tabularx}{\textwidth}{@{}lXccccc@{}}
\toprule
\text{Task} &
\text{Description} &
\text{Output} &
\text{Input} &
\text{Map} &
\text{Centers} &
\text{Instances} \\
\midrule
$t$
  & Identify named entity spans
  & $(T+1)^S$
  & $N$
  & $T$
  & $M$
  & $1$ \\
\addlinespace
$t_1$
  & Identify typed positions
  & $(T+1)^C$
  & $C$
  & $T$
  & $M$
  & $1$ \\
\addlinespace
$t_2$
  & Unitize position
  & $\binom{d+1}{2}$
  & $2d$
  & $0$
  & $0$
  & $M$ \\
\bottomrule
\end{tabularx}}

\caption{
The DAG illustrates a decomposed NER task, and the following table contrasts the dimensions of inferential load for the sub-tasks in the decomposition and the complete task.
Taking the aggregate output space, we have $(T+1)^C + M \cdot \binom{d+1}{2} \leq (T+1)^S$, which we show in \Sref{sec:proof-ner}. We can also observe that the decomposition does not introduce additional centers. With the exception of the input space, the decomposition represents a lower inferential load than the full task.}

\label{fig:example-ner}
\end{figure}

\paragraph{Coreference Resolution} We consider next a typed coreference task $t$ in \autoref{fig:example-coref}, where given a document we extract a set of typed mentions and each pair of mentions is classified based on whether they corefer.
We can decompose the complete task into two sub-tasks: mention identification and antecedent linking. For the first sub-task $t_1$, we identify a set of typed mentions from the document, giving us a sequence of mentions $m_1, m_2, \ldots, m_M$. In sub-task $t_2$, for each mention $m_j$, we identify if there exists an antecedent among the previous mentions $m_1, m_2, \ldots, m_{j-1}$. If not, a new entity is introduced, representing $m_j$. We can observe a dependence between $t_1$ and $t_2$, where mentions can be linked only after they are identified. We iterate through each mention pair, so the aggregate output space of the linking task $t_2$ is $\sum_{j=1}^M j = M(M+1)/2$. For sub-task $t_1$, the space of all possible typed spans is $(T+1)^S$, where $S$ is the set of all possible spans and $T$ is the number of types. We introduce $M$ centers at the mention identification stage, and $k$ centers at the linking stage, where there are $k$ entities.
The output complexity for the complete task, is the product of $\left\{ {M \atop k} \right\}$ (i.e. a Sterling number of the second kind denoting the space of possible linking decisions) and $(T+1)^S$ (i.e. the space of possible typed spans). If we assume at least 10 mentions, more than one coreference cluster, and at least one non-singleton coreference cluster, we can observe in \Sref{sec:proof-coref} that the output space of the decomposition is smaller than that of the complete task. This, in combination with no new centers introduced by decomposition gives us a lower inferential load for the decomposition if the additional processing of the input and map space is not dominant (i.e. $w_i, w_m$ are relatively small).

The value of decomposition for annotation efficiency is also empirically supported by \citet{gandhi-etal-2023-annotating}. 
Through annotation timing experiments, we found mention annotation to be almost twice as fast to manually annotate than full coreference resolution, which is supported in our model of inferential load in that the output complexity for $t_1$ is smaller than $t$ by a factor of $\left\{ {M \atop k} \right\}$. 
We also found that decomposing the coreference task into mention detection and linking during annotation results in a more efficient use of annotator time, which also follows from our model.
\begin{figure}[h]
\centering

\begin{tikzpicture}[node distance=0.9cm, >={Stealth}]
    \node[var] (A2) {$t_1$};
  \node[var, right=1.4cm of A2] (B2) {$t_2$};
  \draw[->] (A2) -- (B2);
  \begin{pgfonlayer}{background}
    \node[plate, fit=(B2)] (inner2) {};
  \end{pgfonlayer}
  \node[font=\footnotesize] at (inner2.south east) [anchor=south east, inner sep=2pt] {$M-1$};
  \begin{pgfonlayer}{background}
    \node[plate, inner sep=10pt, fit=(inner2)] (outer2) {};
  \end{pgfonlayer}
  \node[font=\footnotesize] at (outer2.south east) [anchor=south east, inner sep=2pt] {$M$};
\end{tikzpicture}

\vspace{0.5cm}

{
\footnotesize\begin{tabularx}{\textwidth}{@{}lXccccc@{}}
\toprule
Task &
Description &
Output &
Input &
Map &
Centers &
Instances \\
\midrule
$t$
  & Coreference \newline Resolution
  & $\left\{ {M \atop k} \right\}(T+1)^S$
  & $N$ 
  & $k$
  & $M + k$
  & $1$ \\
\addlinespace
$t_1$
  & Mention \newline Identification
  & $(T+1)^S$
  & $N$ 
  & $0$
  & $M$
  & $1$ \\
\addlinespace
\fbox{$t_2$}
& Link $m_j$ to an \newline antecedent
& $j$
& $1$
& $j-1$
& $\smashoperator{\prod_{m_i:\,i<j}}\!\text{noref}(i,j)$
& $M$ \\
\bottomrule
\end{tabularx}
}
\caption{
The DAG illustrates a decomposed coreference task, and the following table contrasts the dimensions of inferential load for the sub-tasks in the decomposition and the complete task.
Let the function $\text{noref}(i,j) = \mathbb{I}(m_j \text{ does not refer to entity}(m_i))$, where the product indicates that there does not exist any active entities that $m_i$ can link to, and it is necessary to create a new entity. In the decomposition, we first identify all mentions $t_1$, and classify each mention pair based on whether the mentions corefer (i.e. whether mention $m_j$ corefers with an antecedent $m_i$) in $t_2$. A new entity type is created when none of the antecedents corefer with $m_j$, giving us $\sum_i \prod_{m_i: i < j} \mathbb{I}(m_j \text{ does not refer to entity}(m_i)) = k$. Therefore, the decomposition does not introduce new centers. We further observe that the $t_2$ aggregate output space is $M(M+1)/2$, giving us $M(M+1)/2 + (T+1)^S \leq \left\{ {M \atop k} \right\}(T+1)^S$, which we show in \Sref{sec:proof-coref}.  }
\label{fig:example-coref}
\end{figure}
\paragraph{Excessively Decomposed NER} For our last example, we consider NER again, but with an alternate decomposition (\autoref{fig:example-bad-ner}). Given a document with $N$ tokens, we first identify untyped positions in the first sub-task $t_1$. In the second sub-task $t_2$, the $M$ positions are unitized. Finally, we perform $T$ binary type classification tasks for each of the $M$ spans in $t_3$. The output complexity of the decomposition is still bounded by the output complexity of the complete task. However, the map space and input space of the decomposition is much higher. We present this example to illustrate the potential cost incurred when the total number of task instances is inflated by decomposition, resulting in a larger aggregate input and map dimension. For a given annotator, if the cost of processing task input and maintaining the set of map atoms ($w_i, w_m$) is high relative to the cost of output complexity $w_o$, then the complete task should not be decomposed. While we did not empirically test this setting in our work \citep{gandhi-bada-strubell}, it is possible that the additional reading and processing costs incurred from excessive decomposition can outweigh the other dimensions of inferential load, making decomposition a poor use of resources. For example, if a model annotator is equally performant for either the typed position identification task or identifying positions and typing in secession, the incurred cost would be strictly higher for the decomposition.

We can also observe more generally that our model of inferential load penalizes the decomposition of enrichment tasks. In this example, typing as a binary classification task doubles the map space and output space, without reducing the inferential load along any other dimension. The potential cognitive relief that would result from meaningful decomposition of classification tasks with respect to a hierarchical schema is not captured by our model of inferential load. It does not take into account the complexity of tag sets (e.g. higher order, ambiguous concepts), in that tags are not treated as divisible, so the potential reduced cognitive load in classifying units with sub-tags is not considered.
\begin{figure}[h]
\centering
\begin{tikzpicture}[node distance=0.9cm, >={Stealth}]
  \node[var] (A) {$t_1$};
  \node[var, right=of A] (B) {$t_2$};
  \node[var, right=of B] (C) {$t_3$};

  \draw[->] (A) -- (B);
  \draw[->] (B) -- (C);

    \begin{pgfonlayer}{background}
    \node[plate, fit=(C)] (p2) {};
  \end{pgfonlayer}
    \node[font=\footnotesize] at (p2.south east) [anchor=south east, inner sep=2pt] {$T$};
    \begin{pgfonlayer}{background}
        \node[plate, fit=(B)(C)(p2)] (p1) {};
      \end{pgfonlayer}
    \node[font=\footnotesize] at (p1.south east) [anchor=south east, inner sep=2pt] {$M$};
\end{tikzpicture}

\vspace{0.5cm}

{\footnotesize
\begin{tabularx}{\textwidth}{@{}lXccccc@{}}
\toprule
\text{Task} &
\text{Description} &
\text{Output} &
\text{Input} &
\text{Map} &
\text{Centers} &
\text{Instances} \\
\midrule
$t$
  & Identify named entity spans
  & $(T+1)^S$
  & $N$
  & $T$
  & $M$
  & $1$ \\
\addlinespace
$t_1$
  & Identify positions
  & $2^C$
  & $C$
  & $T$
  & $M$
  & $1$ \\
\addlinespace
$t_2$
  & Unitize position
  & $\binom{d+1}{2}$
  & $2d$
  & $0$
  & $0$
  & $M$ \\
  \addlinespace
$t_3$
  & Binary position typing
  & $2$
  & $1$
  & $2$
  & $0$
  & $M\cdot T$ \\
\bottomrule
\end{tabularx}}

\caption{
The DAG illustrates an excessively decomposed NER task, and the following table contrasts the dimensions of inferential load for the sub-tasks in the decomposition and the complete task.
all possible spans. 
Let the context window from which we unitize typed positions be $d$, which is a heuristic for the maximum span length. Although the output complexity of the decomposition is smaller than of the complete task (\Sref{sec:ner-proof-excess}), the input space and map space of the decomposition are both higher than the complete task. For the input dimensionality, we can see $N+ 2d\cdot M + MT \geq  N$, and for the map dimensionality, we can see $2T\cdot M \geq T$.}

\label{fig:example-bad-ner}
\end{figure}
\section{Discussion}\label{sec:discussion}
\subsection{Optimal Task Decomposition}

As we observe in the examples from \Sref{sec:examples}, dimensions of inferential load do not uniformly increase or decrease under decomposition. 
The potential benefit of decomposition is focused in reducing the output space complexity, and this is valuable in cases where annotators are in fact relatively sensitive to this dimension, as we denote with $w_o$. 
We argue that the majority of annotators experience this sensitivity; In \Sref{sec:relation-space-inf-load}, we highlighted previous work demonstrating that both machine and human annotators are sensitive to output space complexity.
In this subsection, we consider features of task decompositions that necessarily reduce the aggregate inferential load, in particular with respect to the dimension of output space complexity. 


First, we observe that \textit{center identification sub-tasks tend to represent source components} in the graphical model representation of task decompositions. 
Enrichment sub-tasks are dependent on center identification sub-tasks, so decompositions that follow the natural flow of annotation would represent center identification as independent sub-tasks.
For enrichment tasks, it is necessary to identify an entity apriori to characterize and enrich with additional information.
Independent source sub-tasks are naturally executed first in the topological sort of decomposition graphs, such that enrichment sub-tasks follow, dependent on the centers identified in the first sub-task(s). 

Second, we can observe that \textit{the mechanism for reducing the inferential load for sub-task instances is center identification}.
Center identification for a sub-task instance will necessarily affect the input space and subsequently the output space complexity for any dependent sub-task instance. 
For any dependency crossing a plate in a graphical model representation of the decomposition, the seed center is passed to the dependent sub-task instance and the attention of the annotator is restricted. 
More formally, for any task incorporating an atom selection primitive, the output space is the atomic map dimension scaled exponentially with the atomic input space dimension. 
For each atom selection sub-task performed to identify centers (e.g. identify positions, identify mentions), the input space shrinks for the following sub-task to the seed centers. 
Accordingly, we argue that the tasks that can be decomposed with lower aggregate inferential loads contain at least one center identification sub-task.

For such tasks, decompositions that reduce the inferential load share the following features:
(1) Center identification is isolated from enrichment, and 
(2) the number of candidate centers is minimized.
The dominating term in the aggregate output space complexity is a function of these two features. This term is the size of the map space scaled exponentially with the size of the input space, where the input space is tied to (2) and the map space is tied to (1).


 

\paragraph{Efficient decompositions isolate center identification sub-task}
In optimal task decompositions, the center identification sub-task is isolated completely from enrichment. 
Since the center identification sub-task(s) represents the source component of the decomposition graph, the output space of center identification is the atomic map dimension scaled exponentially with the atomic input dimension. 
We can expect the center identification task to represent the largest input dimensionality of all the sub-tasks, with the largest resultant output dimensionality. 
In order to reduce the output dimensionality for the source component, the map space can be reduced as much as possible. 
For example, rather than typing and detecting mentions jointly, our model of inferential load would prefer to detect mentions, and type those mentions in secession, as the aggregate output space would be lower for the decomposition if the mention types are not included in the map space at mention detection time.
If we design a decomposition such that center identification sub-tasks are performed first and isolated from other sub-tasks that characterize and enrich the centers, then subsequent sub-tasks operate under a smaller input space and subsequently have a smaller output space. 

\paragraph{Efficient decompositions minimize the number of candidate centers}
In order to drive down the output space complexity of center identification tasks, the input space, or the space of possible centers should be as small as possible. The smallest such space of candidate centers is linear with the input space. Consider for example, the position identification sub-task in the decomposition of NER from \autoref{fig:example-ner}: The space of candidate centers is the set of characters, which is exactly the input space. 
This is in contrast with standard annotation practices for span-bound center identification, where the space of candidate centers is the space of all possible spans in the document.
Considering that center identification tasks contribute the dominant term with respect to output space complexity, representing centers as simply as possible is critical to efficient annotation. 
Given a rudimentary surface form representation of a center, it is cheap to perform subsequent atom selection sub-tasks over a small, constrained search space in order to arrive at the final span-bound representation of a center (e.g. fixed-width unitization).

\subsection{When to Abstain from Decomposition}

At the same time, even if a task containing center identification is decomposed according to these two conditions, there may be settings where the potential benefits in reduced output complexity are outweighed by processing costs or cascading errors that can plague decomposed settings.

\paragraph{High Cost of Information Processing}

In cases where the cost of repeatedly processing input space is high, decomposition results in a larger inferential load. 
The developers of GLiNER2, for example, consider such a setting. GLiNER2 is a small multi-task Information Extraction model with fewer than 500 million parameters designed to run on CPU-hardware and trained to perform, in a single pass, compositions of NER, hierarchical structure extraction, and text classification. \citet{zaratiana-etal-2025-gliner2} inspect an enrichment task, finding that decomposing a 0-shot multi-class classification task to per-label forward passes with a 435M DeBERTa model is both more computationally expensive and underperforms GLiNER2. 
This example, in part, highlights the case where training annotators to perform multiple related sub-tasks jointly may be more efficient than developing specialized annotators for individual sub-tasks.

For annotation of complex structure, however, it is unlikely in practice that the cost of information processing in the decomposed setting is less significant than that of the output complexity in the complete task setting.
For annotation of any text-bound structure, there will be at least one center identification task.
As the number of dependent entities that make up the structure increase, the output space complexity exhibits iterated exponential growth with the number of entities.
Under a decomposition, on the other hand, the cost of processing the map and input space grows polynomially for each task subdivision.
Considering the strong relationship between atomic space complexity and inferential load cited in prior work (\Sref{sec:relation-space-inf-load}), we argue that the cost of information processing in the decomposed task setting will generally be lower than the potential output complexity in the complete task setting for annotators, especially in cases of complex structure with multiple interdependent salient entities.

\paragraph{Cascading Errors under Decomposition} Pipelined annotation systems are necessarily vulnerable to cascading errors \citep{he-etal-2013-dynamic}, and the benefit from modeling related tasks in multi-task learning paradigm may outweigh the additional learning overhead for individual annotators. With respect to model annotation and neural architectures in particular, multi-task learning can improve generalization ability \citep{caruana1997multitask}. For structured annotation tasks such as semantic role labeling \citep{peng-etal-2018-learning}, coreference resolution \citep{swayamdipta-etal-2018-syntactic}, and dependency parsing \citep{peng-etal-2017-deep}, multi-task learning with auxiliary objectives have been shown to improve model performance on individual tasks. The superiority of such systems over pipelined approaches is no longer as universal, however, as finetuned pretrained language models can be robust sub-task annotators, minimizing the effect of cascading errors \citep{gururangan-etal-2020-dont}. At minimum, such model annotators can perform sub-tasks that are downstream from high quality pre-annotation of core, ancestral, sub-structures.

\subsection{Annotator Allocation: The Strongest Annotator Identifies Centers}

In annotation resource constrained settings, concentrating resources on center identification sub-tasks will likely result in higher quality data. 
For one, center identification has a disproportionately large impact on the overall annotation quality of the complete task. Center identification represents the source component of the decomposition graph, so robust annotation of center identification sub-tasks would evade cascading errors in dependent sub-tasks.
Second, we should expect that center identification tasks are cognitively hard and consequently expensive.
This follows from our model of inferential load and centering theory \citep{grosz1995centering}. 
In our model of inferential load, we count the number of new centers introduced for each task, such that center identification tasks have at least one new center while enrichment tasks necessarily have no new centers.
Center identification represents a clear shift in focus, where the salient entity is realized during annotation. 
Such tasks represent a higher cognitive load than enrichment tasks where there is continuity in the center.

\subsection{Interaction between Centering Theory and Atomic Space Complexity}

Our model of inferential load is consistent with centering theory  \citep{grosz-etal-1995-centering} in that tasks that introduce centers, or shift the attention of annotators, have a large atomic output space. On the other hand, tasks that enrich or characterize centers (e.g. type position, unitize position, link mention to an existing center) necessarily have a small atomic output space. 
Such tasks represent a sequence of continuity for annotators, where centers are repeatedly reused. 

The notion that reducing the atomic space can reduce the inferential load is also grounded in centering theory. For smaller input atomic spaces, there are fewer candidate entities or concepts to process while performing annotation, so we can expect annotators to experience fewer shifts in focus. 
Similarly, with respect to output space complexity, if there are fewer potential valid outputs to select between, the annotator experiences fewer shifts in focus when weighing candidate annotations.

\section{Task Distribution Procedure}\label{sec:procedure}
In this section, we propose Algorithm \autoref{algo:assign} for distributing sub-tasks among annotators. We estimate the performance of each annotator for each sub-task on a held-out annotated set, where the cheapest performant model is assigned to each sub-task. 
In the event that such an assignment exceeds the budget $B$, we propose to solve an optimization problem to select a student and teacher annotator for distillation, where we distill a higher quality annotator to a weaker, but cheaper annotator to achieve the best possible performance under the budget constraint.

Sets of heterogeneous annotators can include both a range of models with variable size and a range of human annotators such as crowd-workers or costly domain experts.
The cost of distillation includes both the cost of development of training materials and the cost of training. 
If the teacher annotator is a model, the cost of training material development is the cost of inference to produce silver-annotated data. 
If the teacher annotator is a human, the training material cost includes the time required to annotate examples or develop instructions to correct annotation errors. 
With respect to the cost of training, if the student annotator is a model, distillation follows the standard practice of training the weaker model with annotations produced by the stronger, teacher annotator \citep{hinton2015distilling}. 
For human student annotators, the cost of training includes the time it takes to review annotated examples or instructions.

\begin{algorithm}[t]
\small
\caption{Assign Annotators to Sub-tasks under Budget Constraint
\\
\textbf{Inputs:} Sub-tasks $\{t_1, \ldots, t_{\ell}\}$, annotators $\{A_1, \ldots, A_m\}$, dataset $\mathcal{D}$, quality thresholds $\{Q_1, \ldots, Q_{\ell}\}$, annotation budget $B$\\
\textbf{Output:} Annotation configuration $\texttt{useStrong}, \texttt{useDistilled}$
}
\label{alg:assignment}

\begin{algorithmic}[1]

\Function{assign-annotators}{$\{t_1, \ldots, t_{\ell}\}, \{A_1, \ldots, A_m\}, \{Q_1, \ldots, Q_{\ell}\}, \mathcal{D}, B$}

\For{$i = 1$ to $\ell$}\Comment{Initialize annotation assignments}
    \For{$j = 1$ to $m$}
        \State $\texttt{useStrong}[i][j] \gets \textsc{False}$
        \For{$k = 1$ to $m$}
            \State $\texttt{useDistilled}[i][j][k] \gets \textsc{False}$
        \EndFor
    \EndFor
\EndFor
\State $\mathcal{D}_{\text{dev}} \gets \textsc{Sample}(\mathcal{D})$
\State $\mathcal{D}_{\text{dev}} \gets \textsc{Annotate}(\mathcal{D}_{\text{dev}})$

\For{$A_j \in \{A_1, \ldots, A_m\}$} \Comment{Estimate annotator performance for sub-tasks}
    \For{$t_i \in \{t_1, \ldots, t_{\ell}\}$}
        \State $\hat{Q}_i^j \gets \textsc{Eval}(\mathcal{D}_{\text{dev}}, A_j, t_i)$
    \EndFor
\EndFor

\For{$t_i \in \{t_1, \ldots, t_{\ell}\}$} \Comment{Construct initial greedy assignment}
    \State $j' \gets \arg\min_j\{c_i^j | \forall j \in [m], \hat{Q}_i^j \geq Q_i\}$
    \State $\texttt{useStrong}[i][j] \gets \textsc{true}$
    
\EndFor

\If{\textsc{TotalCost}(\texttt{useStrong}) $\le B$} \Comment{Assignment satisfies budget constraint}
    \State \Return \texttt{useStrong, useDistilled}
\Else \Comment{Optionally select annotator pairs for distillation}

    \State $\texttt{useStrong, useDistilled} \gets \textsc{Distilled-Assignment}(\hat{Q}, Q)$ 
    \State \Return $\texttt{useStrong, useDistilled}$

\EndIf

\EndFunction
\end{algorithmic}
\label{algo:assign}
\end{algorithm}


We describe in this section more formally how to select optimal distillation configurations 
We let $m$ annotators $A_1, A_2, \ldots, A_m$ and performance metrics for each sub-task $i$: $\hat{Q}_i^1, \hat{Q}_i^2, \ldots, \hat{Q}_i^m$.  
The threshold performance for each sub-task is $Q_i$. Assigning annotator $j$ to sub-task $i$ has cost $c_i^j$, and the total budget is $B$.
We denote the performance of annotator $j$ on sub-task $i$ as $\hat{Q}_i^j$.
The performance threshold for sub-task $i$ is denoted by $Q_i$. With respect to cost, $c_i^j$ is the cost of performing sub-task $i$ with annotator $j$, and the distillation cost $d_{i}^{j,k}$ is the sum of the costs of annotating a batch of data with teacher annotator $k$ and training a weak student annotator $j$ on the annotated batch. 
\paragraph{Objective} In Algorithm \autoref{algo:assign}, $\textsc{Distilled-Assignment}$, refers to the following optimization objective (\autoref{eq:objective}) and constraints.
$\textsc{Distilled-Assignment}$ selects, for each sub-task, either a strong annotator or a distilled annotator derived from a stronger annotator, with the goal of maximizing overall annotation quality while satisfying a total budget constraint. The decision variables \texttt{useStrong}$_{i,j}$ and \texttt{useDistilled}$_{i,j,k}$ indicate whether a strong annotator $A_j$ or a distilled annotator derived from $A_j$ to $A_k$ is used for sub-task $i$, respectively.
The objective is to maximize the total expected quality of annotations across all sub-tasks, with both direct annotation with strong annotators and distillation-based annotation with weak annotators. Each sub-task contributes according to the estimated performance of the selected annotator: $\hat{Q}_i^j$ for strong annotators and $\hat{Q}_i^k$ for distilled configurations, where the latter is proxied by the performance of the selected teacher annotator.
\begin{align}
\max \sum_i \Bigl[
\sum_j \texttt{useStrong}_{i,j}\cdot \hat{Q}_i^j
+ \sum_{j,k}\texttt{useDistilled}_{i,j,k}\cdot \hat{Q}_i^k
\Bigr] \label{eq:objective}   
\end{align}

\paragraph{Constraints} The optimization is subject to several constraints. First, exactly one configuration---either a strong annotator or a distillation pair---must be assigned to each sub-task. Second, only annotators whose estimated performance exceeds the sub-task threshold $Q_i$ may be used, including both strong annotators and teachers used in distillation. Third, the feasible set of teacher and student choices is restricted to Pareto-efficient annotators with respect to estimated performance and cost, ensuring that no dominated annotator is considered in the assignment. Fourth, the total cost of all assignments, including both direct annotation cost and the cost of distillation (i.e. teacher annotation and student training), must not exceed the total budget $B$. Finally, all decision variables are binary, reflecting the discrete selection of either a strong annotator or a valid distillation configuration for each sub-task.
\begin{center}
\footnotesize{
\begin{tabular}{@{}p{\linewidth}@{}}
\toprule
Constraints to Satisfy for Annotator Distillation Pair Selection ($\textsc{Distilled-Assignment}$)
\\\midrule
$\displaystyle\sum_j \texttt{useStrong}_{i,j} + \sum_{j,k}\texttt{useDistilled}_{i,j,k} = 1 \quad \forall i$
\hfill \textit{\color{gray} $\triangleright$ One annotator per sub-task} \\[8pt]

$\texttt{useStrong}_{i,j} = 0 \;\; \text{if } \hat{Q}_i^j < Q_i$
\hfill \textit{\color{gray} $\triangleright$ Strong annotators meet threshold} \\[6pt]

$\texttt{useDistilled}_{i,j,k} = 0 \;\; \text{if } \hat{Q}_i^k < Q_i$
\hfill \textit{\color{gray} $\triangleright$ Teacher must meet threshold} \\[6pt]


$\texttt{useDistilled}_{i,j,k} = 0 \;\; \text{if } \exists k' :
\hat{Q}_i^{k'} \ge \hat{Q}_i^{k} \land c_i^{k'} \le c_i^{k}$
\\\hspace{10em}
$\land (\hat{Q}_i^{k'}, c_i^{k'}) \neq (\hat{Q}_i^{k}, c_i^{k})$
\hfill \textit{\color{gray} $\triangleright$ Pareto-efficient teachers only} \\[10pt]

$\texttt{useDistilled}_{i,j,k} = 0 \;\; \text{if } \exists j' :
\hat{Q}_i^{j'} \ge \hat{Q}_i^{j} \land c_i^{j'} \le c_i^{j}$
\hfill \textit{\color{gray} $\triangleright$ Pareto-efficient students only} \\[10pt]

$\displaystyle\sum_i\Bigl[
\sum_j \texttt{useStrong}_{i,j}\cdot c_i^j
+ \sum_{j,k}\texttt{useDistilled}_{i,j,k}\cdot (d_i^{j,k} + c_i^j)
\Bigr]\le B$
\hfill \textit{\color{gray} $\triangleright$ Satisfy budget} \\[10pt]

$\texttt{useStrong}_{i,j},\;\texttt{useDistilled}_{i,j,k}\in\{0,1\}$
\hfill \textit{\color{gray} $\triangleright$ Binary variables} \\

\bottomrule

\end{tabular}}
\end{center}









\section{Related Work}

\paragraph{Modeling Task Complexity}
Prior work conceptualizes task complexity from both model and human-centric perspectives. Model-based approaches characterize complexity in terms of the number of reasoning units required \citep{tang2025importance}, descriptive capability classes \citep{wang-etal-2025-task}, or the size of the smallest model capable of performing the task \citep{bae2023complexitynet}. In parallel, human annotation studies use proxies such as annotation time \citep{Goel2023LLMsAA},  disagreement between expert and lay annotators \citep{yang-etal-2019-predicting}, and multi-dimensional linguistic features correlated with effort \citep{wei2018clinical}. Many of these dimensions can be mapped into our framework—for example, input length corresponds to the atomic input space, while the number of entities reflects the introduction of centers.
Similar to this work, \citet{fort-etal-2012-modeling} propose a broad taxonomy of annotation complexity. 
We differ in explicitly modeling centers as units of annotator focus and in introducing inferential load as a unified measure that can be applied to both human and machine annotators, extending naturally to heterogeneous annotator populations.
In contrast with prior work, we propose a simplified model of inferential load to support redesign of annotation tasks for the sake of efficiency in multi-annotator settings. 

\paragraph{Task Decomposition as a Principle for Resource Efficiency}
Task decomposition for efficient annotation is closely related to partial annotation, where effort is focused on informative sub-structures rather than full structure annotation. This approach has been effective across tasks such as named entity recognition, dependency parsing, event extraction, and relation extraction \citep{zhang-etal-2023-data, marcheggiani-artieres-2014-experimental}. Our work is focused on distributing annotation efforts across sub-tasks, but efficiency could be further improved by granular discrimination of informative sub-task instances with partial annotation.

This perspective aligns with output-constrained annotation and interactive correction, where fixing parts of the structure reduces the search space of remaining decisions. For example, \citet{culotta2005reducing} show that partial corrections can constrain model predictions, effectively converting segmentation into simpler classification problems, highlighting that segmentation complexity grows exponentially with sequence length. Similarly, \citet{michael-2023-case} advocate constructing complex annotations by bootstrapping from simple, narrowly scoped representations. These approaches mirror our notion of centers as core sub-structures that constrain downstream annotation and reduce effective complexity.

\paragraph{Patterns in Annotation Task Decomposition}
Across a range of tasks, decomposition has been adopted to reduce the aggregate inferential load. For example, span extraction is often split into head identification followed by boundary detection, improving robustness in nested NER and mention detection \citep{lin-etal-2019-sequence, peng-etal-2015-joint}. There have been similar decomposition approaches to document-level pseudo-coreference with sub-tasks such as pronoun identification, antecedent resolution, antecedent boundary resolution, antecedent head resolution \citep{jauhar-etal-2015-resolving, liu-etal-2016-exploring}. Conditioning-based approaches fix part of the structure to simplify the remaining structure, as in QA-SRL, which decomposes semantic role labeling into predicate identification and argument detection, enabling efficient crowd-sourced annotation with minimal expertise \citep{he-etal-2015-question}, further optimized through constrained interfaces such as autocomplete \citep{fitzgerald-etal-2018-large}. Related approaches to enable crowd-sourcing decompose complex questions into atomic operations \citep{wolfson-etal-2020-break}. More broadly, effective decomposition enables sub-tasks that can be performed by non-expert annotators or LLMs \citep{michael-2023-case}, consistent with early visions of interactive information extraction systems where users specify behavior through simple examples without requiring domain expertise \citep{cardie1998proposal}.

\section{Conclusion}
In this work, we propose to decompose structured annotation tasks, isolating center identification as a sub-task for more efficient allocation of annotator effort, specifically in cases where aggregate inferential load resulting from decomposition is lower than performing a complete task with a single annotator.
We formalize inferential load as a weighted sum of several dimensions of complexity (\Sref{sec:framework}). This includes the units that are processed and realized over the course of annotation, specifically the atomic input space, map space, and output complexity.
The inferential load incorporates the introduction of centers, or salient discourse entities that are realized in performing annotation, which we borrow from centering theory \citep{grosz1995centering}. 
Through examples in NER and coreference resolution, we illustrate how decomposition of structured annotation tasks, where center identification sub-tasks are isolated and performed first, can reduce the output complexity of subsequent sub-tasks by constraining the focus of annotation (\Sref{sec:examples}). 
Given a decomposition that does reduce aggregate inferential load, we present an algorithm for selecting annotators for each sub-task, where we either greedily assign annotators to sub-tasks or develop annotators that meet budget constraints using distillation for each sub-task (\Sref{sec:procedure}).

We highlight several important limitations of this work.
Our notion of inferential load does not account for the ambiguity of individual atoms or whether the atoms are implicitly compositional in the input or map space which would likely have an effect in practice.
For simplicity, this work also does not distinguish between the space of valid and reasonable annotations.
Decomposition necessarily results in context from the input space processed repeatedly for each sub-task instance, and there may be annotators for whom information processing is especially costly.
Finally, this work presents a theoretical model of efficient decompositions, and while we have demonstrated the value of two particular decompositions for NER and coreference resolution \citep{gandhi-etal-2023-annotating, gandhi-bada-strubell}, we have not explored alternate decompositions for these two tasks, let alone other structured representations.
Despite these limitations, task decomposition is widely applicable for structured annotation under heterogeneous annotator settings, and we present a framework for the efficient design of such decompositions. 

\section*{Acknowledgments}
Thanks to David Mortensen and Sireesh Gururaja for valuable feedback on early drafts of this work.







\bibliographystyle{acl_natbib} 
\bibliography{anthology,custom}
\appendix
\section{NER Atomic Output Space}\label{sec:proof-ner}

We first show that the aggregate output space of the NER decomposition in \autoref{fig:example-ner} is smaller than the output space complexity of the complete task:

\[
(T+1)^S \ge (T+1)^C + M \cdot \binom{d+1}{2}
\]
Taking the logarithm of both sides, we will show that $\log\left((T+1)^S\right) \ge \log\left((T+1)^C + M \cdot \binom{d+1}{2}\right)$.
\small{\begin{align}
\log\Big((T+1)^C + M \cdot \binom{d+1}{2}\Big) 
&= \underbrace{\log\Big((T+1)^C \big(1 + \frac{M \cdot \binom{d+1}{2}}{(T+1)^C}\big)\Big)}_{\text{decompose log product}} \\
&= \log((T+1)^C) + \log\Big(1 + \frac{M \cdot \binom{d+1}{2}}{(T+1)^C}\Big) \\
&= C \log(T+1) + \underbrace{\log\Big(1 + \frac{M \cdot \binom{d+1}{2}}{(T+1)^C}\Big)}_{\log(1+x) \leq x} \\
&\le C \log(T+1) + \underbrace{\frac{M \cdot \binom{d+1}{2}}{(T+1)^C}}_{\text{$M \cdot \binom{d+1}{2} \leq (T+1)^C$}} \\
&\le C \log(T+1) + \underbrace{1}_{\log(T+1)\geq 1}  \\
&= C\log(T+1) + \log(T+1) \\
&= \underbrace{(C+1)}_{ T+1 \leq S}\log(T+1) \\
&\le S \log(T+1) \\
&= \log(T+1)^S 
\end{align}}
In line 7, we assume that the space of $M$ spans ($M \cdot \binom{d+1}{2}$) is smaller than that of all possible typed spans in the document ($(T+1)^C$). We assume on line 10 that the set of possible spans for a document is much larger that the number of types ($T+1 \leq S$).

\section{Coreference Resolution Atomic Output Space}\label{sec:proof-coref}

We show that the aggregate output space of the coreference decomposition is smaller than the output space of the complete task. We assume both that the number of mentions is larger than the number of entities (which is necessarily the case if all the coreference clusters are not singleton) and that there is more than one coreference cluster for a given document, giving us $M >> k \geq 2$. We will prove the following claim.

\[
(1+T)^S \left\{ {M \atop k} \right\} \geq \frac{M(M+1)}{2} + (1+T)^S
\]
First, we can observe that Stirling numbers of the second kind are lower bounded \citep{rennie1969stirling}, so it is enough to show 
\small{\begin{align}
(1+T)^S \left\{ {M \atop k} \right\}\geq (1+T)^S \cdot\frac{1}{2}(k^2+k+2)k^{M-k-1} \geq \frac{M(M+1)}{2} + (1+T)^S \label{eq:stirling-bound}
\end{align}}

Second, we show a supporting claim that $k^{M-k-1} \geq M(M+1)$ for $k\geq 2, M \geq 10$.  $k^{M-k-1}$ is exponential in $M$ as the exponent $M-k-1$ grows linearly with $M$. $M(M+1)$ is quadratic in $M$. Exponential growth dominates polynomial growth, giving us the inequality. We show this more formally.
\small{\begin{align*}
\underbrace{k^{M-k-1}}_{k \geq 2} &\geq 2^{M-k-1}   \\
&\geq 2^{M-3} \\
&= \underbrace{\frac{2^M}{8}}_{M\geq 10} \\
&\geq M(M+1)
\end{align*}}

Finally, starting from the inequality we can then show the claim \autoref{eq:stirling-bound}:
\small{\begin{align}
\underbrace{k^{M-k-1}}_{k\geq 2} &\geq M(M+1) \\
(k^2+k+2)k^{M-k-1}  \geq k^{M-k-1} &\geq M(M+1)\\
\frac{1}{2}(k^2+k+2)k^{M-k-1}  &\geq \frac{1}{2}M(M+1)\\
(1+T)^S\cdot\frac{1}{2}(k^2+k+2)k^{M-k-1}  &\geq \underbrace{(1+T)^S\cdot\frac{1}{2}M(M+1)}_{\text{adding and subtracting }1}\\
&= \frac{M(M+1)}{2} + ((1+T)^S -1)\underbrace{\frac{M(M+1)}{2}}_{\frac{M(M+1)}{2} \geq 1} \\
&\geq \frac{M(M+1)}{2} + ((1+T)^S -1) \\
&\geq \frac{M(M+1)}{2} + (1+T)^S  \\
\end{align}}

\section{Excessively Decomposed NER Atomic Output Space}\label{sec:ner-proof-excess}
To show that the output space of the decomposition is smaller than that of the complete task, it is enough to show the following, in combination with \Sref{sec:proof-ner}.
\[(T+1)^C + M \cdot \binom{d+1}{2} \geq 2^C + M \cdot \binom{d+1}{2} + 2MT\]
Subtracting $M \cdot \binom{d+1}{2}$  and taking the logarithm of both sides, we will show that $\log\left((T+1)^C + \right) \ge \log \left(2^C + 2MT\right)$.
\small{\begin{align}
    \log \left(2^C + 2MT\right) &= \log\left(2^C\left(1+ \frac{2MT}{2^C}\right)\right) \\
    &= \log\left(2^C\right) + \underbrace{\log\left(1+ \frac{2MT}{2^C}\right)}_{\log(1+x) \leq x} \\
    &\leq C\log2 + \underbrace{\frac{2MT}{2^C}}_{2^C> 2MT} \\
    &\leq C\log 2 + \underbrace{1}_{C \geq 6 > \frac{1}{\log\left(3/2\right)}} \\
    &\leq C\log 2 + C \log \left(\frac{3}{2}\right) \\
    &= \underbrace{\log(3)}_{T \geq 2} \\
    &\leq C \log(T+1)
\end{align}}
In line 24, we assume that the space of all possible position assignments is larger than the space of mention type assignments, since the number of position assignments grows exponentially with document length. We also assume at least 6 characters ($C\geq 6$) in the document and 2 types ($T\geq 2$). 
\end{document}